\documentclass[10pt,twocolumn,letterpaper]{article}

\usepackage{wacv}
\usepackage{times}
\usepackage{epsfig}
\usepackage{graphicx}
\usepackage{subfig}
\usepackage{amsmath}
\usepackage{amssymb}
\usepackage{booktabs}
\usepackage{stfloats}
\def\wacvPaperID{****} % Enter the WACV Paper ID here

\wacvapplicationstrack % Uncomment this line if you are submitting to the Applications Track.

\wacvfinalcopy % *** Uncomment this line for the final submission

\ifwacvfinal
\usepackage[breaklinks=true,bookmarks=false]{hyperref}
\else
\usepackage[pagebackref=true,breaklinks=true,colorlinks,bookmarks=false]{hyperref}
\fi

\begin{document}

%%%%%%%%% TITLE
\title{Point Cloud Video Anomaly Detection Based on Point Spatio-Temporal Auto-Encoder}

\author{Tengjiao He\\
School of Electronic and Information Engineering\\
Beihang University, Beijing 100191, China\\
{\tt\small sy2202318@buaa.edu.cn}
\and
Wenguang Wang\\
School of Electronic and Information Engineering\\
Beihang University, Beijing 100191, China\\
{\tt\small wwenguang@buaa.edu.cn}
}

\maketitle
\thispagestyle{empty}

%%%%%%%%% ABSTRACT
\begin{abstract}
   Video anomaly detection has great potential in enhancing safety in the production and monitoring of crucial areas. Currently, most video anomaly detection methods are based on RGB modality, but its redundant semantic information may breach the privacy of residents or patients. The 3D data obtained by depth camera and LiDAR can accurately locate anomalous events in 3D space while preserving human posture and motion information. Identifying individuals through the point cloud is difficult due to its sparsity, which protects personal privacy. In this study, we propose Point Spatio-Temporal Auto-Encoder (PSTAE), an autoencoder framework that uses point cloud videos as input to detect anomalies in point cloud videos. We introduce PSTOp and PSTTransOp to maintain spatial geometric and temporal motion information in point cloud videos. To measure the reconstruction loss of the proposed autoencoder framework, we propose a reconstruction loss measurement strategy based on a shallow feature extractor. Experimental results on the TIMo dataset show that our method outperforms currently representative depth modality-based methods in terms of AUROC and has superior performance in detecting Medical Issue anomalies. These results suggest the potential of point cloud modality in video anomaly detection. Our method sets a new state-of-the-art (SOTA) on the TIMo dataset.
\end{abstract}

%%%%%%%%% BODY TEXT
\section{Introduction}

Video anomaly detection has gained significant attention from the research community in recent years due to its great potential for security surveillance in various settings, including personal dwellings, production, and clinical environments \cite{berroukham2023deep,jiang2022survey, abbas2022comprehensive}. However, the infrequency, diversity, and ambiguity of anomalous events pose great challenges for collecting and customizing a class-balanced dataset for fully-supervised models. As a result, general anomaly detection approaches attempt to learn the data distribution of ``normal" data and classify samples that significantly deviate from this distribution as ``anomaly". Video anomaly detection methods typically use videos containing only normal behaviors to train models, enabling them to learn normal patterns and identify frames with significant deviations as anomaly frames during testing. Such methods are known as unsupervised video anomaly detection (UVAD) \cite{berroukham2023deep}. 

Currently, most video anomaly detection methods are based on RGB modality. Mainly because RGB data acquired cheaply and contains rich color and texture information that enables the extraction of semantic information. However, the use of rich semantic information may also breach personal privacy. Due to the ill-posed and ambiguous nature of depth estimation for 2D data, accurately locating anomalous events in 3D space for RGB images is extremely challenging. The skeleton modality focuses only on human joints and motion information, protecting personal privacy in the scene and remaining insensitive to noise generated by illumination changes. As it does not necessitate the modeling of redundant background changes in the scene, it can focus more on human behavior in the foreground. However, acquiring skeleton data still requires the use of RGB images. And accurately obtaining 3D motion patterns from 2D data remains a challenging task due to the depth blur problem \cite{mishra2022skeletal}.

Point cloud data is usually acquired using depth camera and LiDAR and is presented in 3D. Human point clouds preserve body pose and motion information while protecting personal privacy through their sparsity. Point cloud data enables more accurate positioning of anomalous events in 3D space and avoids the depth blur problem caused by 2D-3D transformation. In contrast to RGB camera, depth camera and LiDAR are insensitive to illumination or color changes, and can be operated 24 hours a day. 

However, there is currently little research on video anomaly detection based on point cloud. Some works, known as ``point cloud anomaly detection", mainly focus on two types of tasks: ``open-set detection" and ``geometric anomaly detection" \cite{wong2020identifying, masuda2021toward, bergmann2021mvtec}. As these tasks are oriented toward object-level anomalies, they deal poorly with the scenario-level anomalies encountered in video anomaly detection \cite{heidecker2021application}. Point cloud video understanding, also called 4D point cloud sequence processing or dynamic point cloud processing, employs static point cloud processing and time series modeling to solve 3D action recognition and 4D semantic segmentation problems \cite{wang20203dv, fan2021deep, zhang2023complete}. While 3D action recognition shares similarities with point cloud video anomaly detection, it differs in that it does not require the model to possess frame-level positioning capability, and it does not conform to the open-set assumption of anomaly detection. Some methods detect anomalous behaviors using point cloud data \cite{iqbal2021detection, ling2022intelligent, nguyen2019estimation}, but they are either limited to a few particular types of anomalies or lack quantitative results on large-scale datasets. This makes it difficult to view them as a counterpart with the video anomaly detection task in the computer vision field.

In this work, we propose an auto-encoder framework, called Point Spatio-Temporal Auto-Encoder (PSTAE), that uses point cloud videos as input and adopts the paradigm of unsupervised video anomaly detection. During training, the auto-encoder reconstructs multiple normal input point cloud frames to learn the patterns of normal behaviors. At test time, the model distinguishes between normal and anomalous frames by outputting a higher reconstruction loss when reconstructing anomalous input point cloud frames. Inspired by the impressive performance of point cloud video understanding methods in 3D action recognition, we introduce the Point 4D Convolution layer (P4Dconv) to process input point cloud videos. Just like what the 3D convolution layer does to RGB videos, P4Dconv can also maintain the spatial geometric information and temporal motion information contained in point cloud videos. The P4Dconv we use in this study are the Point Spatio-Temporal Operation (PSTOp) and Point Spatio-Temporal Transposed Operation (PSTTransOp) proposed by PSTNet++ \cite{fan2021deep}, and that’s why we call our auto-encoder framework Point Spatio-Temporal Auto-Encoder (PSTAE). Auto-encoders designed for RGB videos measure reconstruction loss by comparing the R, G, and B channel values of each pixel in the input and output while keeping the coordinates of each pixel unchanged. However, each point in point cloud videos often contains only coordinate information and lacks channel attributes, making it hard for autoencoders designed for RGB videos to process them directly. To capture the attributes of point cloud videos, local geometric descriptors can be extracted for each point. But for the scenario discussed in this study, the local descriptor extraction networks used for 3D point cloud registration are too voluminous and computationally inefficient \cite{bergmann2023anomaly, deng2018ppf}. To address this problem, we propose a simpler strategy that utilizes a shallow feature extractor to extract local geometric descriptors from raw point cloud videos at a lower computational cost. Several works \cite{mishra2022skeletal, chen2023multiscale, schneider2022unsupervised, khaire2022semi} have shown that focusing more on the foreground can effectively improve the model's anomaly detection performance. Taking advantage of the point cloud data's property of easily separating foreground and background, we apply a background subtraction algorithm to extract the foreground in the scene, allowing PSTAE to focus exclusively on human behavior. Thus we eliminate the noise introduced by background changes and greatly reduce the model's computational cost. 

Currently, two publicly available datasets, TIMo \cite{schneider2022timo} and HARD-ATM \cite{khaire2022semi}, can be used for point cloud video anomaly detection. Both datasets contain depth images that can be converted into point clouds. Studies based on TIMo report more detailed and comprehensive evaluation results focusing only on the depth modality than those based on HARD-ATM. So we chose TIMo for our point cloud video anomaly detection study. 

Our main contributions are as follows: 
\begin{itemize}
\item[$\bullet$] We propose Point Spatio-Temporal Auto-Encoder (PSTAE), an auto-encoder framework based on the Point 4D Convolution layer, to solve the problem of point cloud video anomaly detection. We introduce the PSTOp and PSTTransOp to preserve the spatial geometric information and temporal motion information contained in point cloud videos during encoding and decoding processes. And use an auto-encoder structure to meet the open-set assumption of anomaly detection and enable frame-level anomaly positioning.
\item[$\bullet$] We propose a reconstruction loss measurement strategy that utilizes a shallow feature extractor to help PSTAE to measure reconstruction loss with low computational overhead. The shallow feature extractor is pre-trained to extract local descriptors for each point. The reconstruction loss is then formed by comparing the local descriptors of each point in the input and output of the autoencoder, making the PSTAE trainable.
\item[$\bullet$] Experimental results on the TIMo dataset show that our method achieves a new state-of-the-art (SOTA) performance on that depth video anomaly detection dataset.
\end{itemize}

%------------------------------------------------------------------------
\section{Related Work}

%-------------------------------------------------------------------------
\subsection{Video Anomaly Detection}

Early works relied on object-tracking techniques to model the trajectories of normal targets and defined targets that significantly deviated from normal trajectories as anomalies. The limitation of these methods is that they can only detect trajectory-related anomalous behaviors. To address the diversity of anomalous events and the ambiguity of their definitions, subsequent works focused more on unsupervised pixel-level methods. These methods used hand-crafted features, such as histograms of oriented gradients (HOGs), histograms of optical flow (HOF), and spatio-temporal gradients, or employed sparse coding and dictionary learning.

Due to the excellent performance in computer vision tasks such as object detection and semantic segmentation, an increasing number of deep learning-based methods have emerged in video anomaly detection. With the advent of generative models, recent algorithms have adopted two main styles: reconstruction-based and prediction-based approaches \cite{le2023attention}. The reconstruction-based approaches mainly use auto-encoder-like structures to discriminate normal and anomalous frames by comparing reconstruction errors \cite{chong2017abnormal}, while the prediction-based approaches mainly use GAN-like structures to distinguish frames by comparing prediction errors of future frames \cite{liu2018future}.

More recent works focused on weakly supervised learning, continual learning, fully unsupervised learning, and other areas to better meet the needs of real-world applications \cite{sultani2018real, doshi2022rethinking, zaheer2022generative}. To reduce the labor required to annotate a large number of anomalous frames and increase the amount of usable data, weakly supervised learning algorithms mainly use multiple instance learning (MIL) to distinguish frame-level anomalies from video-level labeled data \cite{sultani2018real}. To continually learn new normal patterns during surveillance and take environmental changes into account, Doshi and Yilmaz \cite{doshi2022rethinking} proposed a continual learning framework for video anomaly detection and a specialized dataset along with a set of evaluation metrics designed for continual video anomaly detection. Current unsupervised video anomaly detection algorithms are actually one-class classifiers trained only with normal data, but due to the powerful generalization ability of DNN, a one-class classifier is also likely to learn an invalid classifier boundary. Although the video-level annotation method of weakly supervised algorithms is less laborious than the frame-level annotation method, given the vast amount of usable video data in the real world, the number of videos that the weakly supervised algorithms can utilize is still quite limited. Therefore, Zaheer et al. \cite{zaheer2022generative} proposed a generative cooperative learning framework and introduced a novel negative learning approach to learn a classifier boundary from completely unlabelled video data, thus achieving true fully unsupervised learning.

Because current anomalous behaviors definitions are often associated with human body posture and the need to protect the privacy of monitored individuals in specific scenarios, video anomaly detection based on skeleton or depth modality has also received a lot of attention from the research community \cite{chen2023multiscale, schneider2022unsupervised, khaire2022semi}. Chen et al. \cite{chen2023multiscale} designed a graph convolutional neural network based on human skeleton key points for video anomaly detection. Human pose estimation algorithms are used to obtain human skeleton key points from the video first. Then following the Deep Embedded Clustering (DEC) framework, Chen conducted three scales of graphs with three adjacency matrices to perform graph convolution in the spatial dimension, And conducted average pooling over multiple time spans to perform key points aggregation and graph convolution in the temporal dimension. As a result, multi-scale representations and attention mechanism are integrated into spatial and temporal dimensions to enhance video anomaly detection performance. Schneider et al. \cite{schneider2022unsupervised} first achieved unsupervised video anomaly detection on depth modality. They trained and tested five types of autoencoder structures commonly used in RGB video anomaly detection on the TIMo depth video anomaly detection dataset and compared the influence of the proportion of foreground loss contained in the loss function on models’ performance. Khaire and Kumar \cite{khaire2022semi} proposed a semi-supervised CNN-BiLSTM autoencoder fusing RGB and depth data, applied in the bank ATM. They first used the pre-trained CNN MobileNet to extract features from each frame of the video segment. Then they concatenated the features obtained from two modalities and finally, a Bi-LSTM-based autoencoder reconstructed the features output by the feature extractor to learn the representations of normal behaviors.

In video anomaly detection, studies on RGB modality are relatively advanced, while other modalities remain to be explored. Skeleton and depth modalities have gradually received attention from the research community due to their unique advantages, and point cloud, which is partially similar to these modalities yet has its own characteristics, has not yet received attention in video anomaly detection. Therefore, we propose this study with the hope of drawing more scholarly attention to this area.

%-------------------------------------------------------------------------
\subsection{Point Cloud Anomaly Detection}

As mentioned in \emph{Section 1 Introduction}, point cloud anomaly detection mainly focuses on two types of tasks: ``open-set detection" and ``geometric anomaly detection". 

Open-set detection methods are utilized to classify targets that fall outside of the pre-defined categories in the training set as ``unknown" and are commonly used to improve object detection robustness in autonomous driving settings. Wong et al. \cite{wong2020identifying} first implemented the idea of point cloud open-set detection by proposing an Open-Set Instance Segmentation (OSIS) network. Regardless of semantic information, point clouds are clustered in the embedding space, enabling the network to classify similar object instances into one cluster without knowing their categories. During inference, the network calculates the feature distances between the input point cloud instance and each cluster in the embedding space and then assigns the instance to a certain cluster. If the distances between the input instance and each existing cluster are too long, the instance will be clustered into a new ``unknown" cluster using the DBSCAN algorithm. Different from Wong's idea of inputting the whole scene, Masuda et al. \cite{masuda2021toward} proposed a variational autoencoder structure based on the idea of FoldingNet \cite{yang2018foldingnet} that takes individual object point clouds as input. It learns normal patterns by reconstructing the geometry of the object point cloud defined as normal through folding operation, and uses the Chamfer distance between the input point cloud and the reconstructed point cloud as anomaly score. Recently, Cen et al. \cite{cen2022open} proposed a Redundancy Classifier (REAL) framework for open-world semantic segmentation tasks. By incorporating multiple redundancy classifiers into the original segmentation network to classify ``unknown" objects, REAL addresses the issue of forgetting former classes that may occur during the incremental learning process. Floris et al. \cite{floris2022composite} proposed the Composite Layer, a point convolution layer that separates spatial encoding and semantic encoding apart. And uses it to build a CompositeNet for 3D anomaly detection. To obtain metrics for measuring anomalies, N kinds of ``geometric transformations" were performed on the original point cloud to form self-supervised labels, allowing CompositeNet to learn features useful for anomaly detection when classifying N kinds of ``geometric transformations". Finally, the average posterior probability of the correct transformations is used as anomaly score. 

Geometric anomaly detection refers to the detection and localization of abnormal protrusions, depressions, defects, and cracks on the surface of the object point cloud. It is frequently employed in industrial quality inspection. Since the MVTec 3D-AD \cite{bergmann2021mvtec} is a recently proposed comprehensive dataset for unsupervised 3D anomaly detection and localization tasks, most recent works are trained and evaluated on this dataset. Bergman and Sattlegger \cite{bergmann2023anomaly} proposed a 3D Student-Teacher network (3D-ST) and designed a self-supervised training strategy to train the teacher network to extract local geometric descriptors from high-resolution point clouds. During training, the student network reconstructs the local geometric descriptors extracted from normal samples. When applied to test data, anomalous structures are detected by regression errors between the output of the student and teacher network. Cao et al. \cite{cao2023complementary} proposed a Complementary Pseudo Multimodal Feature (CPMF) that combines handcrafted feature FPFH with pseudo-2D features to fully utilize local geometric and global semantic information to improve geometric anomaly detection performance. 

According to Heidecker's classification of anomalies \cite{heidecker2021application}, both of the aforementioned types of tasks target object-level anomalies that occur only on a single instance in a single frame. However, the issue we are discussing falls within the category of scenario-level anomalies, which occur on multiple instances in multiple frames.

Several studies have been conducted in point cloud anomaly detection for scenario-level anomalies. Iqbal et al. \cite{iqbal2021detection} proposed a method for detecting anomalous motion around a vehicle using point cloud scene flows acquired by LiDAR. Firstly, scene flows are estimated from adjacent point cloud frames using the proposed Nearest point loss and Cosine-Cycle Consistency loss and embedded in VoxCF representation. Next, a pre-trained 3D CNN extracts motion features from the VoxCF of objects closest to the vehicle. Finally, an LSTM network is trained with motion features of objects exhibiting normal motion, to predict the motion features of the next frame. During testing, the proposed method is expected to produce high prediction errors when applied to objects exhibiting anomalous motion. Ling et al. \cite{ling2022intelligent} proposed an abnormal event detection method applied in rail transit. Human body point clouds acquired by LiDAR are projected onto a 2D plane and transformed into 2D images. The proposed method then obtains the human body silhouette by applying dilation and corrosion on those 2D images, estimates the human body pose using OpenPose algorithms, and dynamically models a five-point inverted pendulum from the human body pose using the Lagrange equation method. Finally, the method analyzes the cumulative force states on the inverted pendulum model to determine if the human body is about to fall. Nguyen and Meunier \cite{nguyen2019estimation} proposed an abnormal gait detection method based on an autoencoder. Depth images are converted to point clouds and encoded as grayscale images using cylindrical histograms. The autoencoder is then trained on grayscale images of normal gait, and the reconstruction error of the autoencoder is used as anomaly score during testing. All the above methods use point clouds to detect anomaly behaviors, but they are either limited to a few specific types of anomalies within limited application scenarios or lack quantifiable metric results on large-scale datasets, while video anomaly detection task often requires a richer variety of anomalies and is typically supported by large amounts of data.

\begin{figure*}
\begin{center}
\includegraphics[width=1.0\linewidth]{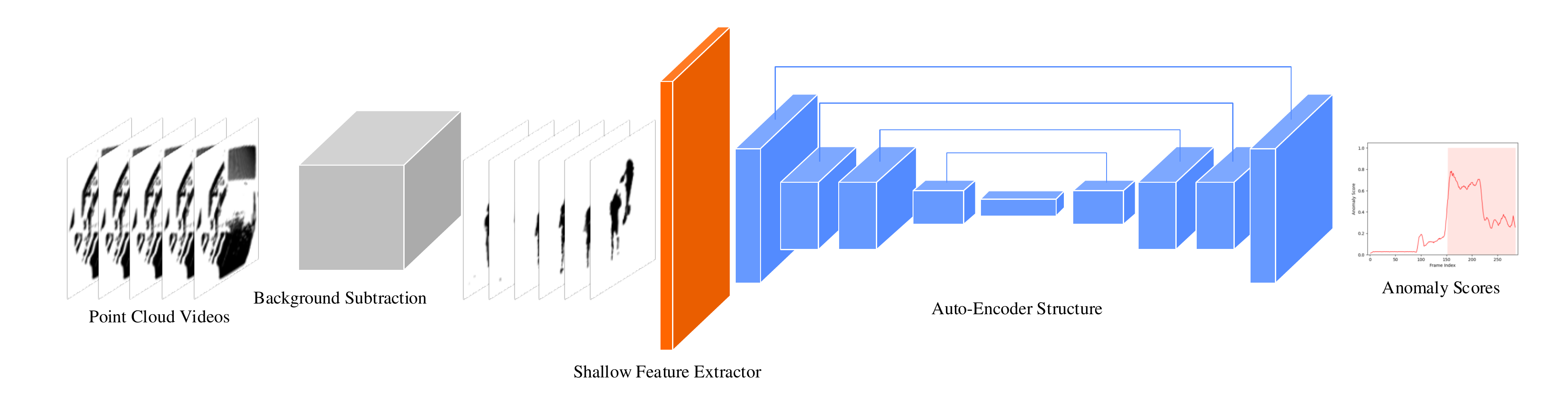}
\end{center}
   \caption{Overall pipeline of our proposed Point Spatio-Temporal Auto-Encoder (PSTAE) framework.}
\label{Figure 1}
\end{figure*}

%-------------------------------------------------------------------------
\subsection{Point Cloud Video Understanding}

Point cloud video, also known as 4D point cloud sequence or dynamic point cloud, encompasses various tasks such as 3D action recognition, 4D semantic segmentation, 3D action segmentation, and scene flow estimation. For the task of 3D action recognition, Wang et al. \cite{wang20203dv} argued that the 3D human pose estimation algorithms relied upon by skeleton-based methods are not robust enough. Furthermore, previous methods that convert depth images into dynamic images and use CNN for processing lose some spatial and motion information. To solve these problems, Wang mapped depth image videos to point cloud videos, voxelized the point cloud videos into voxel set videos, compressed the voxel set videos via temporal rank pooling into one voxel set (3DV), and finally used PointNet++ \cite{qi2017pointnet++} to process the voxel set to recognize actions. However, voxelizing the point cloud can lead to quantization errors and costly computational expenses. Therefore, Liu et al. \cite{liu2019meteornet} proposed MeteorNet, which directly processes point cloud videos. They aggregate features of the adjacent points around the center point through the Meteor module and stack Meteor modules to obtain global spatiotemporal contextual features. Fan et al. \cite{fan2022pstnet, fan2021deep} found the idea of tracking points to obtain regional dynamics, as proposed by MeteorNet, somewhat difficult to implement. Moreover, simply connecting spatial and temporal dimensions ignores the temporal regularity and the scale difference between spatial and temporal domains. Therefore, they proposed PSTNet \cite{fan2022pstnet} and PSTNet++ \cite{fan2021deep} to model spatio-temporal local structures without relying on point tracking through the proposed ``point tube". They reduce the impact of the spatial irregularity of points on temporal modeling by decoupling spatial convolution and temporal convolution. 

Recent efforts aim to introduce transformers to enhance the extraction of spatio-temporal information \cite{wei2022spatial, fan2022point, wen2022point}, and to overcome the problem of limited annotations in point cloud videos through self-supervised \cite{zhang2023complete} and semi-supervised learning \cite{chen2022maple}.

Point cloud 3D action recognition is very similar to point cloud video anomaly detection. Both tasks involve encoding temporal information and identifying human behaviors. However, 3D action recognition does not require the model to possess frame-level positioning capability, and the categories of actions the model can recognize are predefined, which does not conform to the open-set assumption of anomaly detection.

%------------------------------------------------------------------------
\section{Point Spatio-temporal Auto-Encoder}

Our proposed Point Spatio-Temporal Auto-Encoder (PSTAE) framework comprises a background subtraction module, an auto-encoder built with PSTOp and PSTTransOp, and a shallow feature extractor. Each input point cloud video is divided into nonoverlapping segments of fixed length in the time dimension. And PSTAE takes those segments as input to detect frames with anomalies. Figure \ref{Figure 1} shows the overall pipeline of our method. Each component is discussed in the rest of this chapter.

%-------------------------------------------------------------------------
\subsection{Background Subtraction}

In surveillance, anomalies are more likely to occur in the foreground than in the almost unchanged background. But most video anomaly detection methods do not differentiate between foreground and background when processing, leading to redundant modeling of the background. Additionally, clutter in the background and noise caused by illumination changes can cause false alarms. To address these issues, some methods use human pose estimation algorithms, attention mechanisms, or pre-trained object detectors to focus on foreground changes. Without complex calculations, background subtraction can be easily applied to point cloud videos to completely remove the influence of background and focus solely on foreground changes since point cloud videos are captured from a fixed perspective and presented in 3D. For RGB or depth videos with a fixed number of pixels per frame, background subtraction would not significantly affect the computational cost of the model. But for Point 4D Convolution layers where computational cost is positively related to the number of points in each frame, background subtraction can greatly improve the model's computational efficiency.

In this work, We applied a simple yet effective point cloud background subtraction algorithm. First, we voxelize the point cloud by dividing the space into grids with length, width, and height equal to $r$. Then, we calculate the cumulative point density $D$ of each grid over consecutive $l$ frames, set a point density threshold $\theta $, and classify the points in the grid with $D>\theta $ as background point cloud and the points in the grid with $D\le \theta $ as foreground point cloud. The insight behind this is that the background point cloud is considered static, while the foreground point cloud is relatively dynamic. By removing points in the background voxels, we get the foreground point cloud videos.

%-------------------------------------------------------------------------
\subsection{Auto-Encoder Structure}

Auto-encoder is a widely used unsupervised learning architecture, which is often designed as a cascade of an encoder and a decoder. The encoder compresses the input into a representation in latent space, while the decoder reconstructs the original input data from the latent representation. When training the auto-encoder to reconstruct the input data, the auto-encoder captures the dominant distribution of the input data in the latent space.

To maintain the spatial and temporal information contained in RGB videos when applying the auto-encoder principle to RGB video anomaly detection, CNN, ConvLSTM, and Transformer are often used instead of Fully Connected (FC) layers to construct the autoencoder. Then, a training set consisting only of normal videos is fed to the autoencoder. During the testing phase, while normal frames are expected to have a lower reconstruction loss when reconstructed by the autoencoder, anomalous frames will have a higher reconstruction loss due to their greater distance from the dominant distribution in the latent space. The reconstruction loss can subsequently be used as anomaly score.

Similar to using convolutional and deconvolutional layers to construct the Convolutional Auto-Encoder (CAE), we introduce the Point 4D Convolution layer to maintain the spatial geometric information and temporal motion information in point cloud videos during reconstruction. Specifically, we use PSTOp and PSTTransOp proposed in PSTNet++ \cite{fan2021deep} to construct the autoencoder, and name our autoencoder framework as Point Spatio-Temporal Autoencoder (PSTAE).

Similar to the 3D convolution kernel, PSTNet++ proposes the ``Point Tube". 3D convolution kernel aggregates features in the spatiotemporal neighborhood on image sequences. Just like what the 3D convolution kernel does on RGB videos, PSTOp uses the farthest point sampling (FPS) algorithm to obtain anchor points and aggregates the features of the points in the spatial neighborhood around each anchor point. Next, frames obtained by downsampling with a fixed stride in the temporal dimension are used as anchor frames, and the features of the points in the temporal neighborhood around each anchor frame are aggregated. PSTTransOp uses skip connections to get the original coordinates of the points from the corresponding PSTOp layer and propagates the features of anchor points to the corresponding spatiotemporal neighborhoods. More detailed description and principle of the Point Spatio-Temporal Operation can be referred to in \cite{fan2021deep}.

%-------------------------------------------------------------------------
\subsection{Reconstruction Loss Measurement Strategy}

Each pixel in RGB videos has its fixed coordinates and R, G, and B channels. This makes it convenient for an auto-encoder to measure reconstruction loss by comparing the values of the three channels on each pixel between the input and reconstructed output.

For the convolutional autoencoder used for RGB video anomaly detection, the reconstruction loss is often designed as mean squared error (MSE). As shown in Equation (\ref{Equation (1)}), where $N$ is the number of pixels in the input image, ${{x}_{i}}\in {{R}^{d}}$ represent the color values of the $i$-th pixel in the input image, $d=3$ for RGB images, and $\widehat{{{x}_{i}}}\in {{R}^{d}}$ represent the reconstruction values of the $i$-th pixel output by the convolutional autoencoder.
\begin{eqnarray}
{{L}_{rgb}}=\frac{1}{N}\mathop{\sum }_{i=1}^{N}{{\left( {{x}_{i}}-\widehat{{{x}_{i}}} \right)}^{2}}
\label{Equation (1)}
\end{eqnarray}

In contrast to RGB videos, point cloud videos typically lack attributes corresponding to each point beyond coordinate information. This makes measuring reconstruction loss using autoencoders based on the Point 4D Convolution layer challenging. To solve this problem, local geometric descriptors can be extracted for each point. But directly introducing a local descriptor extraction network designed for 3D point cloud registration is computationally expensive and inefficient for the problem at hand. To address this, we propose a simpler strategy for measuring the reconstruction loss by utilizing a pre-trained shallow feature extractor to extract local geometric descriptors.

The shallow feature extractor is a PSTOp layer that does not perform temporal down-sampling or aggregate motion information. It is called ``shallow" because it consists of only one layer. According to the proof in \cite{fan2021deep}, such a PSTOp layer is equal to PointNet++ \cite{qi2017pointnet++}. Given a point cloud video segment $\mathcal{P}=\left( {{P}_{1}},{{P}_{2}},...,{{P}_{L}} \right)$ with a length of $L$, where ${{P}_{t}}\in {{R}^{M\times 3}}$ represents a point cloud frame at time $t$ containing $M$ points, and each point is represented by 3D coordinates $\left( x,y,z \right)$. The point cloud video segment is input to the shallow feature extractor to obtain another segment $\mathcal{P}\text{ }\!\!'\!\!\text{ }=\left( \left[ {{P}_{1}}',{{F}_{1}} \right],\left[ {{P}_{2}}',{{F}_{2}} \right],...,\left[ {{P}_{L}}',{{F}_{L}} \right] \right)$ with the same length. Here, ${{P}_{t}}'\in {{R}^{\frac{M}{s}\times 3}}$ represents an anchor-based point cloud frame after feature extraction, ${{F}_{t}}\in {{R}^{\frac{M}{s}\times f}}$ represents the local descriptor corresponding to each anchor point after feature extraction, $s$ is the spatial downsampling rate of the FPS algorithm, and $f$ denotes the dimension of the extracted local descriptors. The pretraining of the shallow feature extractor can be realized by making it the input layer of a PSTOp-based 3D action recognition network. By training the network on a corresponding dataset, the pretraining of the shallow feature extractor is done at the same time.
\begin{table}
  \begin{center}
    {\small{
\begin{tabular}{ccc}
\toprule
Aggressive Behavior & Medical Issue & Left-Behind Objects \\
\midrule
COF-ARG & COLLAPSE & STP-LBO\\
CROSS-ARG & CRAWL & WALK-LBO\\
PASS-SO & FLOOR & PASS-1LBO\\
RUN-THO & STAGGER & PASS-2LBO\\
WALK-THO & & RUN-LBO\\
CROSS-SO & & COF-LBO\\
\bottomrule
\end{tabular}
}}
\end{center}
\caption{Anomaly categories for each anomalous behavior in TIMo dataset.}
\label{Table 1}
\end{table}

\begin{table}
  \begin{center}
    {\small{
\begin{tabular}{cccccccc}
\toprule
 & ${{r}_{s}}$ & ${{s}_{s}}$ & ${{c}_{s}}$ & ${{r}_{t}}$ & ${{s}_{t}}$ & ${{c}_{t}}$ & ${{p}_{t}}$\\
\midrule
Extractor & ${{r}_{0}}$	& 2	& 4	& 0	& 1	& 8	& [0,0]\\
\midrule
Encoder2 & 2${{r}_{0}}$	& 2	& 45 & 1 & 2 & 64 & [0,0]\\
Encoder3 & 2${{r}_{0}}$ & 1	& 128 & 1 & 1 & 256 & [1,1]\\
Encoder4 & 4${{r}_{0}}$	& 2	& 384 & 1 & 2 & 512 & [0,0]\\
Encoder5 & 8${{r}_{0}}$	& 2	& 768 & 1 & 1 & 1024 & [1,1]\\
\midrule
Decoder5 & - & - & 512 & 1 & 1 & 768 & [-1,-1]\\
Decoder4 & - & - & 256 & 1 & 2 & 384 & [0,0]\\
Decoder3 & - & - & 64 & 1 & 1 & 128 & [-1,-1]\\
Decoder2 & - & - & 8 & 1 & 2 & 45 & [0,0]\\
\bottomrule
\end{tabular}
}}
\end{center}
\caption{Hyperparameters of each layer of the autoencoder structure and pre-trained shallow feature extractor.}
\label{Table 2}
\end{table}

%-------------------------------------------------------------------------
\subsection{Anomaly Scoring}

Given point cloud video clips processed by the shallow feature extractor as input, the reconstruction loss of PSTAE is given by Equation (\ref{Equation (2)}), where $L$ is the length of the input point cloud video clip, ${{F}_{i}}$ represents the local descriptors of the $i$-th frame of the input clip, $\widehat{{{F}_{i}}}$ represents the reconstruction values of PSTAE for the local descriptors of the $i$-th frame, and ${\left\|\cdot\right\|_{F}}$ represents the Frobenius Norm.
\begin{eqnarray}
{{L}_{PSTAE}}=\frac{1}{L}\mathop{\sum }_{i=1}^{L}\left\|{{F}_{i}}-\widehat{{{F}_{i}}}\right\|_{F}^{2}
\label{Equation (2)}
\end{eqnarray}

To obtain the anomaly score, we normalize the reconstruction loss of all frames in each test video to [0,1] and consider that the closer the value is to 1, the higher the probability of an anomalous event occurring. On the other hand, we follow the post-processing for anomaly scores used in \cite{schneider2022unsupervised}. we employ a moving average operation with a window length of 10 to smooth the anomaly scores, making it more robust.
\begin{table*}
  \begin{center}
    {\small{
\begin{tabular}{ccccc}
\toprule
 & Aggressive Behavior & Medical Issue & Left-Behind Objects	& Total\\
\midrule
CAE	& -	& -	& -	& 66.4\\
ConvLSTM & - & - & - & 62.8\\
R-CAE & 82.3 & 52.3	& 68.6 & 70.0\\
P-CAE & 85.9 & 65.3	& {\bf 79.1} & 78.1\\
R-ViT-AE & 79.3	& 60.3 & 73.7 & 71.7\\
P-ViT-AE & 79.3	& 61.8 & 72.6 & 71.2\\
P-ConvLSTM & 78.6 & 52.6 & 66.7	& 67.5\\
\midrule
PSTAE & {\bf 88.6} & {\bf 91.0}	& 75.1 & {\bf 81.0}\\
\bottomrule
\end{tabular}
}}
\end{center}
\caption{Frame-level AUROC (\%) for each method on TIMo Tilted View subset. The best performing method on each anomaly category is highlighted in bold}
\label{Table 3}
\end{table*}

\begin{table*}
  \begin{center}
    {\small{
\begin{tabular}{ccccc}
\toprule
 & Aggressive Behavior & Medical Issue & Left-Behind Objects	& Total\\
 \midrule
CAE	& -	& -	& -	& 56.4\\
ConvLSTM	& -	& -	& -	& 62.2\\
R-CAE	& 86.9	& 75.6	& 73.2	& 73.2\\
P-CAE	& {\bf 93.0}	& 73.8	& 65.1	& 67.8\\
R-ViT-AE	& 91.1	& 68.4	& 64.3	& 65.3\\
P-ViT-AE	& 92.5	& 65.9	& 62.1	& 63.0\\
P-ConvLSTM	& 91.6	& 67.1	& 65.5	& 65.6\\
\midrule
PSTAE	& 86.1	& {\bf 81.7}	& {\bf 75.9}	& {\bf 78.2}\\
\bottomrule
\end{tabular}
}}
\end{center}
\caption{Frame-level AUROC (\%) for each method on TIMo Top-Down View subset. The best performing method on each anomaly category is highlighted in bold}
\label{Table 4}
\end{table*}

%------------------------------------------------------------------------
\section{Experiments}

In this section, we provide details about our experimental setup first. Then we compare our PSTAE with existing SOTA methods. The influence of different components and some qualitative results are also included.

%-------------------------------------------------------------------------
\subsection{Experiment Setup}

We choose to evaluate our method on the TIMo \cite{schneider2022timo} dataset, a large-scale public dataset for unsupervised depth video anomaly detection. TIMo contains depth and infrared videos of choreographic normal and anomalous human behaviors captured by Microsoft Azure Kinect ToF cameras in two indoor scenes. One scene camera captures an open office area with a tilted view and a narrow field of view, while the other scene camera captures an area allowing crossing in all directions at the height of 2.25m, 2.5m, and 2.75m, with a top-down view and a wide FOV. Each video in TIMo only performs one type of normal or anomalous behavior, and Table \ref{Table 1} shows the corresponding anomaly category for each anomalous behavior given in \cite{schneider2022unsupervised}. The training set of TIMo consists of 909 videos (285 from Tilted View and 624 from Top-Down View), comprising 365,979 frames of normal behavior, while the test set comprises 679 videos (182 from Tilted View and 497 from Top-Down View), comprising 245,818 frames, including 110 normal behavior videos and 569 anomalous behavior videos.

We use the camera intrinsics provided by the TIMo dataset to convert depth videos into point cloud videos and remove those invalid points contained in each frame.

Following the existing methods in video anomaly detection study, we use the area under ROC curve (AUROC) for evaluation and comparison. Calculate it using the frame-level anomaly labels provided by the TIMo dataset.

%-------------------------------------------------------------------------
\subsection{Implementation Details}

For background subtraction, we set the grid size $r=0.05$, the accumulated frame number $l=30$, and the point density threshold $\theta =100$. The foreground point cloud videos after background subtraction processing are normalized to 2048 points per frame (sufficient to include all foreground objects in the scene), and the segment size $L$ for each video is set to 15 frames. The hyperparameters of each layer of the autoencoder structure and pre-trained shallow feature extractor are shown in Table \ref{Table 2}, where ${{r}_{s}}$ represents the spatial radius of ball query, ${{s}_{s}}$ represents the spatial subsampling rate, ${{c}_{s}}$ represents the channel number of output in spatial MLP, ${{r}_{t}}$ represents the temporal radius of the temporal neighborhood window, ${{s}_{t}}$ represents the temporal stride, ${{c}_{t}}$ represents the channel number of output in temporal MLP, and ${{p}_{t}}$ denote the beginning padding and the ending padding. The initial radius ${{r}_{0}}$ of the ball query is set to 0.5, and the maximum number of neighboring points in the ball query is set to 9. The pre-training of the shallow feature extractor is performed on the MSR-Action3D \cite{li2010action} dataset. The dimension of the local descriptors $f$ is set to (4,8,16,32) for comparison, and the detailed experimental results can be found in \emph{Subsection 4.3 Experiment Results}. The learnable parameters of the shallow feature extractor are frozen during the training of the autoencoder structure. We train our model for 15 epochs with the SGD optimizer. The batch size is set to 8 and learning rate is set to 0.01, decays with a rate of 0.1 at the 10th epoch. All experiments are performed on an NVIDIA RTX 3080.

%-------------------------------------------------------------------------
\subsection{Experiment Results}

We compare the best performance of our PSTAE with several existing depth-based video anomaly detection methods and baselines proposed by the TIMo dataset, as shown in Tables \ref{Table 3} and \ref{Table 4}. Note that we also follow the work \cite{schneider2022unsupervised} to report results on each anomaly category. (The optimal result of our method is achieved when the local descriptor dimension $f=8$, which will be discussed in the next paragraphs.) The proposed PSTAE outperforms the previous state-of-the-art method on AUROC by about 3\% on the Tilted View subset and about 5\% on the Top-Down View subset, demonstrating the potential of point cloud modality in video anomaly detection. Our PSTAE outperforms the depth-based methods on the anomaly categories of Aggressive Behavior and Medical Issue on the Tilted View subset, and Left-Behind Object and Medical Issue on the Top-Down View subset. Particularly, our PSTAE increases the AUROC of Medical Issue on Tilted View subset by about 25.7\% compared with depth-based methods. In addition, we also report the results on more fine-grained anomaly classification in \emph{Section A Supplement} Tables \ref{Table 5} and \ref{Table 6}. We evaluate the computational efficiency and memory usage of PSTAE. On an NVIDIA RTX 3080, the inference speed of PSTAE without background subtraction module is 176 FPS, and the number of parameters is 7.45M.

\begin{figure}
\begin{center}
\includegraphics[width=1.0\linewidth]{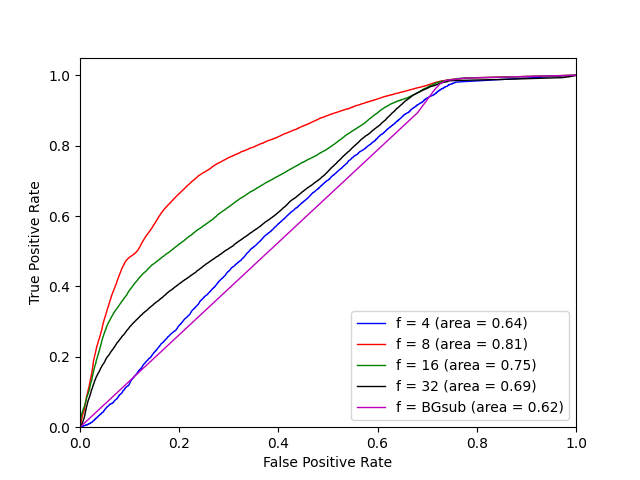}
\end{center}
   \caption{ROC curve and AUC on the Tilted View subset with the local descriptor dimensions f of (4, 8, 16, 32). And a background subtraction baseline (BGsub) guided only by background subtraction. Best viewed in color.}
\label{Figure 2}
\end{figure}

To investigate the effect of the output dimension of the shallow feature extractor on the performance of PSTAE, Figure \ref{Figure 2} shows the ROC curve and AUC on the Tilted View subset with the local descriptor dimensions f of (4, 8, 16, 32). And the optimal performance of PSTAE is achieved with the local descriptor dimension of 8. The possible reason is that when f is too small, the local descriptor may not fully capture the local geometry around the anchor point, whereas when f is too large, the autoencoder may reconstruct redundant dimensions, making the reconstruction loss difficult to converge. We also investigate the influence of background subtraction on the performance of PSTAE by providing a baseline method (BGsub) that relies only on background subtraction to guide the change of anomaly scores, as shown in Figure \ref{Figure 2}. Specifically, the anomaly score of frames with foreground is set to 1 and the score of frames without foreground is set to 0. Even for the worst-performing model whose f equals 4, our PSTAE performs slightly better than the baseline, indicating that the good performance of PSTAE is not solely attributed to background subtraction.

\begin{figure}
\centering
\subfloat[]{
\includegraphics[width=0.32\linewidth]{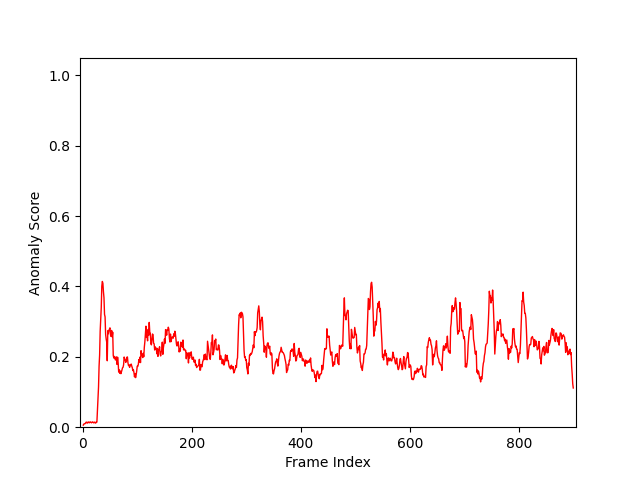}
\label{subfig a}
}
\subfloat[]{
\includegraphics[width=0.32\linewidth]{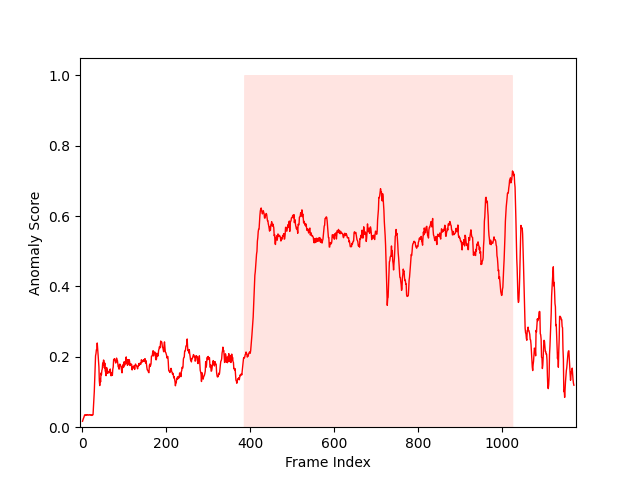}
\label{subfig b}
}
\subfloat[]{
\includegraphics[width=0.32\linewidth]{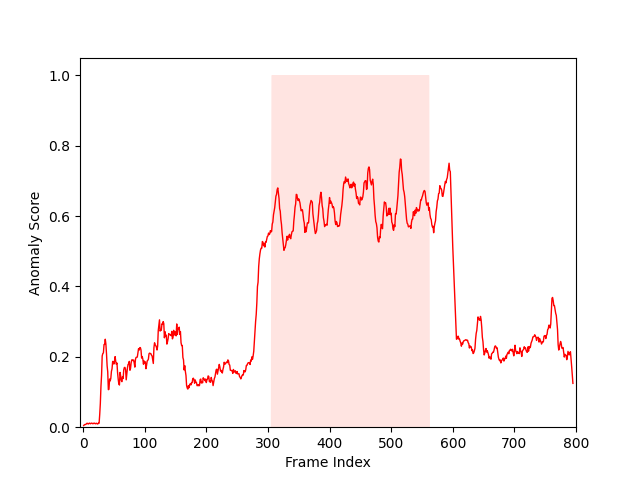}
\label{subfig c}
}

\subfloat[]{
\includegraphics[width=0.3\linewidth]{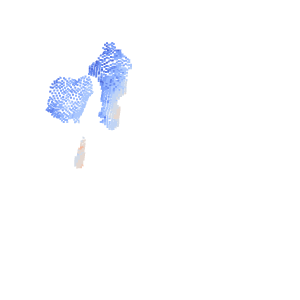}
\label{subfig d}
}
\subfloat[]{
\includegraphics[width=0.3\linewidth]{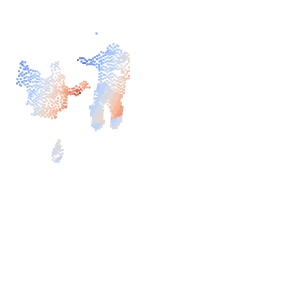}
\label{subfig e}
}
\subfloat[]{
\includegraphics[width=0.3\linewidth]{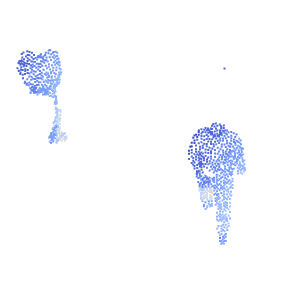}
\label{subfig f}
}
   \caption{Visualization of anomaly scores and point cloud heat maps. (a), (b) and (c) are anomaly scores on three videos in the TIMo test set vary with frames. (d), (e) and (f) are point cloud heat maps depict the reconstruction errors in the scenes of two people meeting, arguing, and parting in video (c). Best viewed in color.}
\label{Figure 3}
\end{figure}

Figure \ref{subfig a}, \ref{subfig b} and \ref{subfig c} illustrate how anomaly scores output by PSTAE on three videos in the TIMo test set vary with frames respectively. Note that the anomaly scores drawn as red lines rise sharply within the range of frames marked as pink (ground truth of anomalies). Figure \ref{subfig d}, \ref{subfig e} and \ref{subfig f} show the point cloud heat maps, which depict the reconstruction errors between the inputs and outputs of PSTAE in the scenes of two people meeting, arguing, and parting in video \ref{subfig c}. The warmer colors of the point cloud heat maps correspond to higher reconstruction errors. Notice that the reconstruction error of the limb parts increases dramatically when the two people act aggressively, indicating that our model effectively detects anomalies.

%------------------------------------------------------------------------
\section{Conclusion}

We propose a simple yet effective point cloud video anomaly detection framework, PSTAE. First, we introduce the background subtraction algorithm to process the original point cloud video. This improves the model's computational efficiency and enables it to focus entirely on foreground changes. Then, we use a shallow feature extractor to extract local geometric descriptors from the input point cloud video to help subsequent autoencoder structure measure reconstruction loss with low computational costs. Finally, we input the point cloud video with local descriptors into the constructed autoencoder structure to achieve frame-level anomaly position under the paradigm of unsupervised video anomaly detection. To preserve the spatial geometric and temporal motion information contained in the point cloud video during the encoding and decoding process of the autoencoder, we introduce the Point 4D Convolution layer to construct the autoencoder. Experimental results on the TIMo dataset show that our method surpasses representative depth-based methods, including P-CAE and R-CAE, in terms of AUROC. Specifically, it is 3\% higher on the Tilted View subset and 5\% higher on the Top-Down View subset. Moreover, our method has significant advantages in detecting Medical Issue anomalies compared to representative methods, achieving a 25.7\% higher AUROC on the Tilted View subset. These results demonstrate the potential of point cloud modality in video anomaly detection. At present, the performance of point cloud video anomaly detection still lags far behind that of RGB video anomaly detection. However, due to its advantages of privacy protection, accurate 3D positioning, and insensitivity to illumination, we believe that this field will receive more attention. We hope that our work can inspire future work and serve as a baseline.

{\small
\bibliographystyle{ieee_fullname}
\bibliography{egbib}
}

\appendix
\newpage
%------------------------------------------------------------------------
\section{Supplement}
\begin{table*}[b]
  \begin{center}
    {\small{
\begin{tabular}{ccccccccccc}
\toprule
	& \rotatebox{90}{COF-ARG}	& \rotatebox{90}{COF-LBO}	& \rotatebox{90}{COF-NONS}	& \rotatebox{90}{COLLAPSE}	& \rotatebox{90}{CRAWL}	& \rotatebox{90}{CROSS-ARG}	& \rotatebox{90}{FLOOR}	& \rotatebox{90}{PASS-2LBO}	& \rotatebox{90}{PASS-SO}	& \rotatebox{90}{RUN-LBO}\\
\midrule
R-CAE	& {\bf 97.4}	& 71.9	& 87.4	& 54.5	& 39.7	& {\bf 90.5}	& 51.0	& 81.3	& 87.2	& 63.9\\
P-CAE	& 96.2	& {\bf 75.5}	& 86.2	& 61.7	& 47.5	& 87.9	& 65.6	& 85.8	& 86.3	& 71.9\\
R-ViT-AE	& 90.4	& 64.2	& 84.7	& 63.7	& 57.2	& 82.6	& 63.6	& {\bf 87.1}	& 84.5	& 67.8\\
P-ViT-AE	& 90.6	& 64.4	& 85.8	& 62.9	& 56.1	& 79.4	& 64.3	& 86.4	& 84.7	& 71.0\\
P-ConvLSTM	& 93.3	& 64.3	& 79.5	& 53.5	& 52.0	& 79.5	& 49.4	& 78.1	& {\bf 88.2}	& 75.2\\
\midrule
PSTAE	& 96.3	& 73.2	& {\bf 98.3}	& {\bf 94.4}	& {\bf 93.6}	& 87.6	& {\bf 92.7}	& 83.1	& 87.6	& {\bf 90.5}\\
\midrule\midrule
	& \rotatebox{90}{RUN-THO}	& \rotatebox{90}{RUN-WEAP}	& \rotatebox{90}{SPORT}	& \rotatebox{90}{STAGGER}	& \rotatebox{90}{STP-LBO}	& \rotatebox{90}{UMB}	& \rotatebox{90}{WALK-LBO}	& \rotatebox{90}{WALK-THO}	& \rotatebox{90}{WALK-WEAP}\\
\midrule
R-CAE	& 79.7	& 68.3	& 61.1	& 75.3	& 75.3	& 64.9	& 64.6	& 67.2	& 60.3\\
P-CAE	& 84.4	& 69.6	& 66.5	& {\bf 87.2}	& 83.2	& 66.0	& {\bf 79.1}	& 80.6	& 57.5\\
R-ViT-AE	& 82.9	& 63.3	& 75.5	& 71.2	& {\bf 87.0}	& 62.8	& 78.9	& 75.7	& 69.3\\
P-ViT-AE	& 83.8	& 66.7	& 71.0	& 71.0	& 86.6	& 64.7	& 78.2	& 76.2	& {\bf 69.4}\\
P-ConvLSTM	& 81.5	& {\bf 84.1}	& 71.5	& 73.1	& 67.2	& 66.6	& 61.5	& 74.5	& 67.2\\
\midrule
PSTAE	& {\bf 89.7}	& {\bf 84.1}	& {\bf 93.3}	& 77.0	& 76.7	& {\bf 75.2}	& 72.6	& {\bf 88.3}	& 57.9\\
\bottomrule
\end{tabular}
}}
\end{center}
\caption{Frame-level AUROC (\%) on more fine-grained anomaly classification on TIMo Tilted View subset. The best performing method on each anomaly classification is highlighted in bold}
\label{Table 5}
\end{table*}

\begin{table*}[b]
  \begin{center}
    {\small{
\begin{tabular}{ccccccc}
\toprule
	& COLLAPSE	& CRAWL	& CROSS-ARG	& CROSS-SO	& FLOOR	& PASS-1LBO\\
\midrule
R-CAE	& 87.6	& 70.8	& 84.0	& 88.9	& {\bf 83.6}	& {\bf 83.1}\\
P-CAE	& 86.5	& 77.0	& {\bf 94.6}	& 92.5	& 78.7	& 74.9\\
R-ViT-AE	& 68.4	& 68.2	& 90.7	& 90.9	& 76.0	& 74.1\\
P-ViT-AE	& 60.1	& 68.0	& 92.5	& 92.5	& 72.0	& 71.8\\
P-ConvLSTM	& 61.3	& 67.7	& 90.7	& {\bf 93.3}	& 74.9	& 76.1\\
\midrule
PSTAE	& {\bf 94.0}	& {\bf 82.2}	& 85.5	& 91.1	& 81.0	& 59.8\\
\midrule\midrule
	& PASS-2LBO	& PASS-SO	& RUN-LBO	& STAGGER	& STP-LBO	& WALK-LBO\\
\midrule
R-CAE	& {\bf 83.7}	& 87.9	& 62.0	& 60.4	& 66.8	& 70.4\\
P-CAE	& 78.5	& 92.0	& 60.8	& 67.3	& 59.2	& 61.3\\
R-ViT-AE	& 79.1	& 91.7	& 58.1	& 63.9	& 56.5	& 59.0\\
P-ViT-AE	& 76.6	& {\bf 92.6}	& 58.7	& 68.1	& 54.5	& 56.6\\
P-ConvLSTM	& 78.4	& 90.8	& 59.0	& 65.6	& 55.9	& 59.9\\
\midrule
PSTAE	& 71.8	& 86.3	& {\bf 83.3}	& {\bf 70.8}	& {\bf 75.0}	& {\bf 83.5}\\
\bottomrule
\end{tabular}
}}
\end{center}
\caption{Frame-level AUROC (\%) on more fine-grained anomaly classification on TIMo Top-Down View subset. The best performing method on each anomaly classification is highlighted in bold}
\label{Table 6}
\end{table*}

\end{document}